# Boosting DNN Cold Inference on Devices

Rongjie Yi[1], Ting Cao[2], Ao Zhou[1], Xiao Ma[1], Shangguang Wang[1], Mengwei Xu[1]

[1]State Key Laboratory of Networking and Switching Technology, Beijing, China

[2]Microsoft Research

{yirongjie,aozhou,maxiao18,sgwang,mwx}@bupt.edu.cn

ting.cao@microsoft.com

## ABSTRACT

DNNs are ubiquitous on edge devices nowadays. With its increasing importance and use cases, it's not likely to pack all DNNs into device memory and expect that each inference has been warmed up. Therefore, *cold inference*, the process to read, initialize, and execute a DNN model, is becoming commonplace and its performance is urgently demanded to be optimized. To this end, we present NNV12, the first on-device inference engine optimizing cold inference. NNV12 is built atop three novel optimization knobs: selecting a proper kernel (i.e., operator implementation) for each DNN operator, bypassing the weights transformation process by caching the post-transformed weights on disk, and pipelined execution of many kernels on asymmetric processors. To tackle with the huge search space, NNV12 employs a heuristic-based scheme to obtain a near-optimal kernel scheduling plan. We fully implement a prototype of NNV12 and evaluate its performance across extensive experiments. It shows that NNV12 achieves up to 15.2× speedup compared to the state-of-the-art DNN engines on edge CPUs and 401.5× speedup on edge GPUs, respectively.

## 1 INTRODUCTION

Deep Neural Networks (DNNs) have become indispensable for mobile applications [55, 70]. Pursuing low inference delay and data privacy, DNN deployment is shifting from large data centers to humble edge devices, e.g., smartphones, IoT, wearables, and autonomous vehicles [22]. Recent work [13, 63] show the number of DNN-embedded apps on Google Play was doubled from Fed. 2020 to Apr. 2021. Those apps have been downloaded by billions of times by users. In essence, *almost every mobile app is becoming a DNN app*.

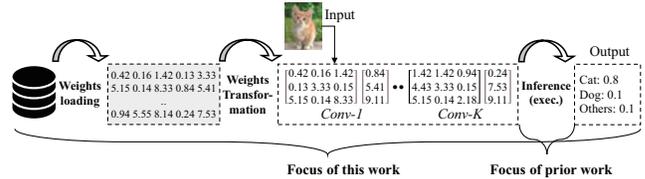

**Figure 1: The process of cold inference.**

Deploying DNNs on devices brings two challenges unpresented on datacenters. One is the tightly constrained hardware resources (memory, compute, energy, etc). In respond, our community has invested tremendous amount of researches on it [23, 34, 38, 56, 57, 59, 61, 62, 64, 67, 69, 72]. Especially, it's necessary to obtain more accurate DNN inference results on devices with limited memory. The second one is the volatile, multi-tenant(app) runtime environment [23, 46], which fundamentally differs from datacenters who typically host a single DNN service on dedicated, highly scalable GPU clusters [71]. It's inevitable to switch between multiple DNN inference on devices with limited memory. Those characteristics lead to a phenomenon that DNNs cannot always reside in device memory; consequently, the DNN inference often occurs in a *cold* manner, i.e., the device needs to load and initialize the model weights into memory before execution.

In general, on-device DNN cold inference could occur in both active and passive manners.

- *Active cold inference* happens per developers' willingness. By design, developers often deliberately avoids a model residing in memory for a long time to reduce memory footprint. For example, certain mobile apps always re-launch DNNs that are infrequently used to reduce its memory usage and thus the probability being killed after moved to background. We observe many such cases from the Google Play apps: PDF scanner [4] and its optical character recognition (OCR) model; image editing and beauty camera apps [1, 5] and their many DNNs as image filters; etc. On intelligent IoTs like Home Hubs [2], cameras [66] and robots [18], DNNs multitasking imposes high pressure on memory as well. The common approach is to pack all DNNs into device memory through weights sharing [19, 21, 38, 39] to avoid cold inference. Those methods, however, are not scalable as with more DNNs the model accuracy drops significantly.

- *Passive cold inference* happens out of the control of developers. This is especially the case for smartphones, where the OS aggressively kills background apps (thus the DNNs) to reduce memory footprint [44, 47]. For instance, a study on 96 Android users show that the app cold launch probability is more than 40% under various background activity scheduling algorithms [37]. Even if memory permits, cold inference still occurs in abundant cases: mobile browsers need to relaunch a DNN whenever certain web pages are opened such as language translation [43]; a DNN-based software could crash and needs to fast re-launch such as in autonomous driving [60]; etc.

In either circumstance, DNN cold inference performance is crucial to user experience and application quality. Its importance is just the same as app launch speed [31, 68] or web page load time (PLT) [49, 58] – two well established and explored research domains by our community. For DNNs that execute in one shot, e.g., the PDF scanner and beauty cameras mentioned above, each inference is cold and its delay directly connects to the application performance. For continuous inference, the delay of the very first inference sometimes can not be simply amortized to the whole inference session. For instance, an auto-driving vehicle or robot need to cold-start an obstacle detection model fast to avoid accidents, either after the model is cleared from memory intentionally or the DL execution engine crashes unintentionally [60].

**Cold inference is poorly supported** Unfortunately, the state-of-the-art DNN engines including `TFLite` [10] and `ncnn` [8] are not ready to boost cold inference as fast as warm inference. As will be shown in §2, the cold inference latency of those engines is 1.5×–12.7× and 85.5×–443.5× slower than warm inference on embedded CPU and GPU, respectively. Taking a step closer, we find the major portions of cold inference include reading the weights from disk into memory (*weights reading*), converting raw weights into an inference-ready format (*weights transformation*), and the actual *model execution*, as shown in Figure 1. Those complicated `operations` distinguish cold inference from traditional warm inference and compromise or even fully invalidate existing techniques in optimizing the inference speed.

For the first time, we propose a system engine, namely NNV12, which *directly* optimizes the DNN cold inference latency on edge devices. NNV12 does not rely on any assumptions of model structures and incurs zero accuracy loss.

**Optimization knobs** (§3.1) We first thoroughly explore the design spaces of cold inference and identify three effective optimization knobs that are rarely touched on in previous literature. (1) *The selection of kernel.* DNN engines typically incorporate many different implementations for one single operator (namely *kernel*), e.g., 28 for convolution in `ncnn`. Those built-in kernels are to improve the inference speed under specific operator configurations, and the current kernel selection is purely based on warm inference speed. However, we observe that the fastest kernel in warm inference does not necessarily exhibit the best performance in cold inference, e.g., a winograd-based kernel [36] executes fast but spends much time in weights transformation. (2) *Post-transformed weights caching.* Weights transformation can be bypassed by storing the post-transformed weights on disk so they can be directly read and executed. However, the transformed weights might occupy more storage and incur higher I/O time. Reading raw or post-transformed weights opens tradeoff among disk I/O and computations. (3) *The order of operator execution and core binding.* The weights reading, transformation, and execution can be pipelined to reduce the blocking time of disk/memory I/O. The pipeline can also orchestrate with the asymmetric processor on edge devices, e.g., CPU/GPU and big.LITTLE core, which can hardly be fully utilized in executing DNNs sequentially.

**Formulation and challenges** (§3.2) The above optimizations need to be jointly considered because their impacts on cold inference are tightly coupled, e.g., choosing a different kernel could overturn an optimal pipeline strategy. To design a holistic and judicious cold inference scheme, we face two primary challenges. First, the search space is too large. We formulate the problem in combined kernel selection, transformation bypassing, and execution pipeline to obtain an optimal kernel scheduling plan. The problem turns out to be NP-hard. Second, the placement of different `operations` (reading, transformation, and execution) could interfere with each other due to the limited disk/memory I/O bandwidth, which further complicates the problem.

**A heuristic-based kernel scheduler** (§3.3) It is inspired by two observations. (1) There exists operation-processor affinity, e.g., the big vs. little core acceleration ratio is more significant for kernel execution than weights reading and transformation. Therefore, the kernel execution is always prioritized on the stronger processor. (2) Multithreading on multiple cores is more efficient on execution `operation` than others. Hence, we only multithread execution `operation` while scheduling other `operations` separately. It also exploits the opportunity that weights reading/transformation `operations` have fewer dependencies than execution, therefore can be easily scheduled individually.

Atop those heuristics, we design an intuitive yet effective kernel scheduling algorithm. Its key idea is to balance the workloads on different processors or cores to minimize the total running time. Meanwhile, during the decision making, NNV12 keeps calibrating the per-operation performance through re-profiling for better scheduling planning. We then extend the above design to the GPU platform (§3.4) by introducing new GPU-specific `operations` into the scheduling pipeline and a shader caching technique. NNV12 also incorporates a workload stealing technique to adapt to dynamic load that could share the hardware resources with cold inference. Furthermore, to ensure that the kernel selection for cold inference does not compromise the warm inference latency in continuous inference (§3.5), NNV12 leverages the spare CPU time slots to switch the kernels.

We've implemented a prototype of NNV12 atop `ncnn` that fully realizes the above techniques. We then perform extensive experiments to evaluate NNV12's performance through 12 typical DNN models and 4 devices including 2 smartphones (CPU) and 2 Jetson embedded devices (GPU). The results show that, on Meizu 16T CPU, NNV12 can reduce the cold inference latency by 5.1×/9.5×/3.7× at average compared to `ncnn`, `tflite`, and AsyMo [56], respectively. On Jetson TX2 GPU, the improvement is even up to 58.2×/401.5× compared to `ncnn` and `TensorFlow`, respectively. NNV12 also greatly reduces the energy consumption of cold inference. The ablation study further shows that each individual technique of NNV12 contributes to significant improvements.

The major contributions of this work are:

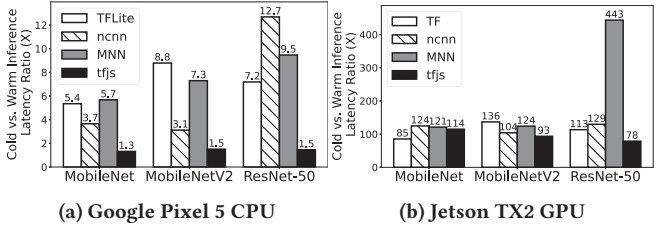

**Figure 2: Cold inference is significantly slower (1.5×−443.5×) than warm inference on vanilla DL libraries.**

- We highlight the importance of NN cold inference and reveal the unsatisfactory support for cold inference quantitatively on the state-of-the-art DNN engines.
- We identify three optimization knobs that can effectively reduce the cold inference latency yet are underexplored in prior literature: kernel selection, weights transformation bypassing, and pipelined inference.
- We propose a holistic framework NNV12 that judiciously considers the three above optimization knobs through a heuristic-based kernel scheduling.
- We implement a prototype of NNV12 and demonstrate its effectiveness through extensive experiments. The code is publicly available at
  https://github.com/UbiquitousLearning/NNV12.

## 2 UNDERSTANDING NN COLD START

Undoubtedly, AI models are going to be prevalent on edge devices. Given the large quantity, a considerable number of model invocations will be cold due to the memory bound, especially for those not frequently used. Indeed we have inspected a few Google Play apps. There are some concrete evidences: (1) The voice assistant such as Siri needs to process the received audio in a cold inference manner each time it is awakened. (2) PDF scanners [4] perform OCR in a cold inference manner. The application runs in a cold inference manner every time it scans images. (3) The beauty camera [1, 5] also employs cold inference. These applications load filters by cold inference when a new filter is selected to beautify a image.

We first perform a set of measurement studies to understand cold inference on edge devices. We use two typical devices: Google Pixel 5 smartphone with Kryo 475 CPU [7] and Jetson TX2 with NVIDIA Pascal GPU [3]. We experiment with 3 DNN models (MobileNetv1/v2, ResNet-50) on 3 popular DL libraries: `TFLite/TF/tfjs`, ncnn [8], and MNN [30].

Figure 2 illustrates the performance gap between cold and warm inference on the above hardware and libraries. As observed, the gap is 1.5×−12.7× on CPU and 85.5×−433.5× on GPU. Concretely, the cold inference latency of ResNet-50 on Kryo 475 CPU takes at least 511.67 ms, while the warm inference only takes 141.56 ms. Such huge gap can inevitably hurt the user experience under scenarios as described in §1.

**Cold inference breakdown** We then investigate the cold inference process internally. While different DL libraries differ in implementation, conceptually their cold inference mainly includes the following stages:

| Device Platform | Google Pixel 5 | Jetson TX2 |
|---|---|---|
| Processor | CPU | GPU |
| **Weights reading** | 36.52 ms | 43.03 ms |
| **Memory allocation** | 1.34 ms | 0.69 ms |
| **GPU preparation** | - | 3004.01 ms |
| **Weights transformation** | 1135.28 ms | 1616.84 ms |
| **Model execution** | 190.12 ms | 802.77 ms |
| **Total cold inference** | 1363.23 ms | 5467.48 ms |
| **Warm inference** | 185.82 ms | 137.02 ms |

**Table 1: A breakdown of ResNet-50 cold inference latency on edge CPU and GPU.**

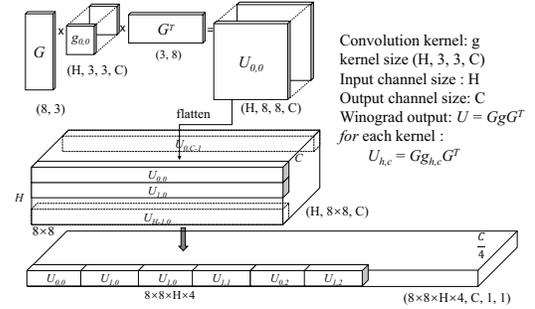

**Figure 3: The weights transformation example for a winograd-based convolution kernel [36].**

- **Memory allocation**: requesting memory from OS to hold the weights and intermediate results during inference.
- **Weights reading**: reading the model weights from device storage into memory.
- **Weights transformation**: converting raw weights into the proper format to facilitate the inference. This process depends on the kernel implementation for each operator. For example, as shown in Figure 3, in a winograd-based convolution kernel [36], the weights will be transformed from size $(H, 3, 3, C)$ to $(8 \times 8 \times H \times 4, \frac{C}{4}, 1, 1)$.
- **Model execution**: the actual inference (forward) process by invoking each operator of the model.
- **GPU preparation** (only for GPU): setting up the GPU driver, creating data pipeline, compiling shader codes, etc.

Table 1 shows the breakdown of cold inference with ResNet-50. On both CPU and GPU, each stage except memory allocation contributes to a considerable portion of the slow cold inference. To obtain an acceptable cold inference latency, we need to optimize each of the above stages.

**Opportunities** Our design is inspired by two key observations. First, DNNs typically have a layer-by-layer computation pattern. As such, the system does not need to wait for the whole model to be loaded to begin the inference execution. Instead, the loading, transformation, and execution of different layers can be possibly pipelined. The concept of layer in DNNs also provides an easy-to-use basis to schedule the I/O, data-intensive, and computation-intensive stages in devices. Second, common DL libraries often provide multiple kernels (i.e., the concrete implementation of operators) for each operator. Those kernels provide a large room of trade-off in disk I/O, memory I/O, and computing.

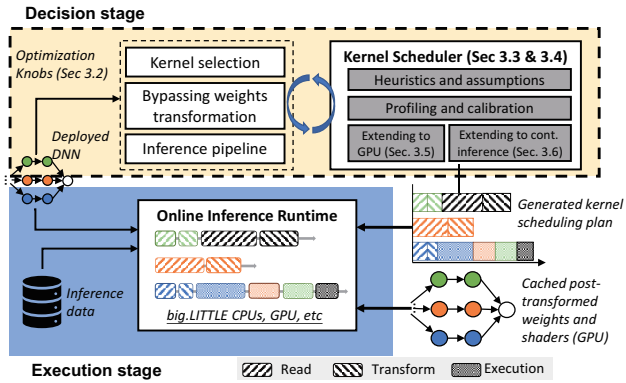

**Figure 4: The simplified workflow of NNV12.**

## 3 NNV12 DESIGN

NNV12 is designed to enable fast NN cold inference on edge devices with the following principles:

- It shall sacrifice zero prediction accuracy.
- It shall require minimal additional efforts from developers.
- It builds atop DL kernels that are already existing in DL libraries. Generating better-optimized kernels is not our contribution and is orthogonal to this work.

The workflow of NNV12 consists of two main stages as shown in Figure 4: offline decision generation and online cold inference runtime. The decision stage is to generate an optimal kernel scheduling plan for a huge design space as explored in §3.1. This stage runs fully automatically on device for one shot, e.g., when a model is fetched to the device, so the decision is optimized for different devices' hardware capacity and requires no efforts from developers. NNV12 follows the generated plan to optimize the cold inference at runtime. From developers' perspective, NNV12 is akin to traditional DL inference libraries in deployment.

### 3.1 Optimization Knobs

We first discuss the optimization knobs that can be explored: kernel selection, weights transformation bypassing, and inference pipeline. While being intuitive, those optimizations have been rarely touched in prior work.

*3.1.1 Kernel selection.* A DNN model can be represented as a directed data graph consisting of many operators. An operator describes how the input data is mapped to output data at a high level, while the kernels represent the different implementation of an operator.

**One operator, many kernels** A key observation we make from existing DL libraries is that there are often multiple kernels implemented for one operator, especially those computation-intensive ones. For instance, as shown in Figure 5, ncnn implements 28 different kernels for convolutional operator.

There are three main reasons for such phenomenon. (1) Kernels can be better optimized with assumptions on the input/weights configurations, e.g., the convolution kernel size and input/output channel numbers. (2) The relative kernel performance relies on the specific hardware platforms. Therefore, developers write multiple kernels to obtain good performance on different platforms by choosing the best fitting. (3) New kernels are emerging but the old ones are kept in the codebase for legacy reasons.

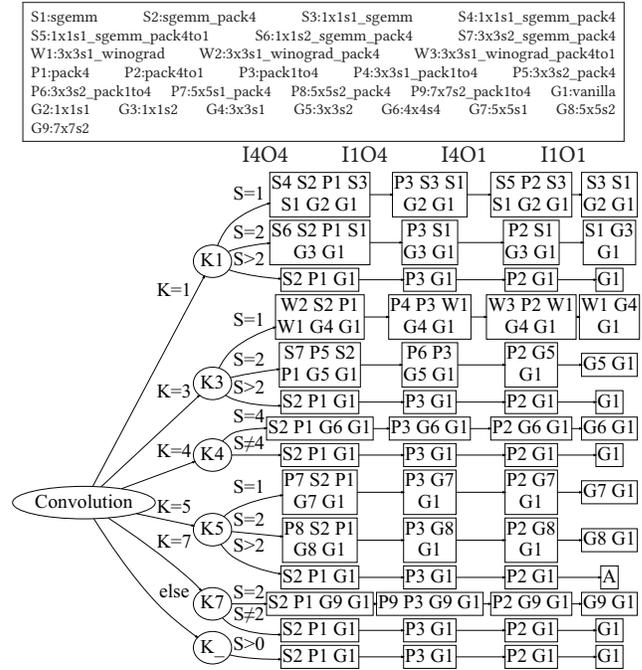

**Figure 5: The 28 kernels implemented by ncnn for convolution. On the top box, "A:B" indicates "B" is a kernel implementation and abbreviated as "A". Within the tree structure, each node (as rectangle) contains the usable kernels for each situation. K/S indicate the convolutional kernel size and stride size. "I4O4" means the input and output channels are divisible by 4.**

**No silver-bullet kernel** The current kernel selection policy of popular DNN engines is hard-coded and only considers the warm inference speed. However, such selection may not be optimal for cold inference. Taking ncnn as an example, as we quantitatively show in Table 2, the default kernel used by ncnn for convolution operators with 4x input/output channel numbers and 3x3 convolution kernel size is a winograd-based implementation (3x3s1-winograd-pack4) because it achieves the fastest warm inference. Such a kernel, however, incurs a high time cost in weights transformation. Instead, a more generic sgemm-based implementation (sgemm-pack4) has less total time cost with simpler weights transformation.

*3.1.2 Bypassed weights transformation.* As shown above, the weights transformation for certain kernels can be extremely costly, despite that the kernel executes quite fast. One possible method to avoid the heavy transformation while leveraging the kernel's fast execution is to cache the transformed weights on disk, which can be directly loaded and executed.

By using this method, NNV12 will cache the post-transformation weights to disk during the decision stage shown in Figure 4, and load these cached weights from disk during the execution stage. This method eliminates the weights transformation time in cold inference, however, could also introduce additional disk storage and I/O.

| Kernels | Cold Inference Time (ms) | | | |
|---|---|---|---|---|
| | Read Raw | Weights Trans. | Read Cache | Execution |
| 3x3s1-winograd-pack4 | 0.70 | 38.23 | 5.23 | 2.98 |
| sgemm-pack4 | 0.70 | 2.21 | 0.70 | 8.14 |
| pack4 | 0.70 | 2.22 | 0.70 | 18.63 |
| 3x3s1-winograd | 0.70 | 65.67 | 4.12 | 3.37 |
| 3x3s1 | 0.70 | 0.00 | 0.70 | 8.01 |
| general | 0.70 | 0.00 | 0.70 | 87.12 |

**Table 2: The weights transformation (on CPU little cores) and execution time (on CPU big cores) of different kernel alternatives for conv op (kernel size = 3, stride = 1, input/output size = 64/192). "Read Raw/Cache" is the I/O time of reading the weights w/o and with cache policy (i.e., pre-transformed).**

As shown in the column "Read Cache" column of Table 2, the post-transformation weights often occupy more storage because the weights will be duplicated. In other words, caching post-transformation weights trades off disk read with memory-access-intensive weights transformation for a given kernel. From the perspective of a whole model's cold inference that consists of many kernels, it opens rich trade-offs between the I/O and memory access.

*3.1.3 Pipelined inference.* Nowadays edge devices are typically equipped with multi-process/core architecture such as big.LITTLE CPU cores. To fully exploit those processors to boost cold inference, one might simply multithread the kernel preparation and execution, e.g., using many threads to read and transform the weights simultaneously. However, we observe the benefits from such multithreading are limited due to two reasons. First, weights reading and transformation stages are not bounded by the computation but disk I/O and memory I/O, respectively. Second, the asymmetric multiprocessor on edge devices makes it difficult to partition the DL workloads to fully utilize each processor's capacity [56], therefore a straggler processor could significantly slow down the whole inference regarding the data flow dependency.

Instead of simply multithreading the kernels separately, we propose to pipeline them: overlap different kernels' weights reading, transformation, and execution. This is based on a key opportunity that DNNs typically have a layer-by-layer computation pattern. As such, the system does not need to wait for the whole model to be loaded or all weights to be transformed to begin the kernel execution. Instead, the loading, transformation, and execution of different layers can be possibly pipelined. The concept of the layer in DNNs also provides an easy-to-use basis to schedule the I/O, memory-intensive, and computation-intensive stages in devices.

## 3.2 Problem Formulation

**The need for a kernel scheduler** To fully harness the optimization knobs introduced in §3.1, we need a global kernel scheduler to determine (i) which kernel to use for each operator; (ii) whether to load the raw weights or the cached post-transformed weights for each kernel; (iii) when and where to execute each `operation`. In this work, we use the term `operation` to indicate each stage of a kernel, e.g., its weights reading, transformation, and execution are three different `operations`. Apparently, those knobs need to be jointly considered as they inherently are coupled with each other.

| Notation | Description |
|---|---|
| $M_l$, $M_b$ | Number of little and big CPU cores |
| $N$ | Number of model layers |
| $r_i$, $w_i$, $e_i$ | The $i$-th read/transform/execution operation ($1 \le i \le N$) |
| $f(i, j, t)$ | Whether operation $i$ executes on core $j$ at time $t$ |
| $S_{i,j}$ | The timestamp when operation $i$ starts running on core $j$ |
| $E_i$ | The timestamp when operation $i$ finish running |
| $T(f(i, j, t))$ | Latency of operation $i$ runs |
| $\Theta_i$ | The set of precursor operations of operation $i$ |
| $\eta$, $F$ | The set of all operations and all inference operations |

**Table 3: Notations used in §3.2 and §3.3.**

For instance, choosing a different kernel could overturn an optimal pipeline strategy.

**Formulation of the kernel scheduling problem** For simplicity, we first use big.LITTLE CPU architecture as the target scheduling platform to introduce our formulation and scheduling scheme. They can be easily extended to other heterogeneous processors, e.g., CPU + GPU as will be discussed later. The annotations used in Tabel 3. We use $f(i, j, t) = 1$ to indicate operation $i$ executes on core $j$ at time $t$ (otherwise 0). Based on that, $S_{i,j}$ and $E_i$ can be expressed by $f(i, j, t)$ in Equation (1). Here, $\eta_i$ means the set of cores where `operation` $i$ runs on.

$$
\begin{aligned}
\eta_i &= \{j | \sum_t f(i, j, t) \ge 1\} \\
S_{i,j} &= \arg\min_t \{f(i, j, t) = 1\}, \quad j \in \eta_i \\
E_i &= \max\{S_{i,j} + T(f(i, j, t))\}, \quad j \in \eta_i
\end{aligned}
\tag{1}
$$

Minimizing the cold-inference latency equals to minimizing the finishing time of the last execution `operation` $f_N$:

$$
\min E_{e_N}
$$

$$
s.t. \begin{cases}
S_{i,j} \ge E_\alpha, & \alpha \in \Theta_i, \forall i, \forall j \\
\sum_{i \in \eta} f(i, j, t) \le 1, & \forall t, \forall j \\
\sum_{i \in \eta} \sum_{j=0}^{M_l + M_b} f(i, j, t) \le M_l + M_b, & \forall t
\end{cases}
\tag{2}
$$

The solver is restricted by three conditions: (1) For each `operation`, its starting time is no earlier than the end time of its all precursor `operations`. We can build a dependency graph among the total $3 \times N$ operations in a DNN by integrating the original dependency of the model (among execution operations) and the read-transform-execution flow of every single kernel. (2) For each core, only one `operation` can run at a given timestamp; (3) At any time, the total number of cores being used should be no larger than $M_l + M_b$.

**Challenges** Solving the above challenges faces the following primary challenges. First, according to Equation (1), $S_{i,j}$ and $E_i$ are nonlinear functions of the optimization variables $f(i, j, t)$. Therefore it is Nonlinear Integer Programming, a classical NP-hard problem. Second, we observe that the execution time $T(f(i, j, t))$ can be interfered with by each other even though they run on different cores. This is mainly because the co-running `operations` reach the limit of disk and/or memory I/O speed. In summary, it's not likely to obtain an optimal kernel scheduling plan directly.

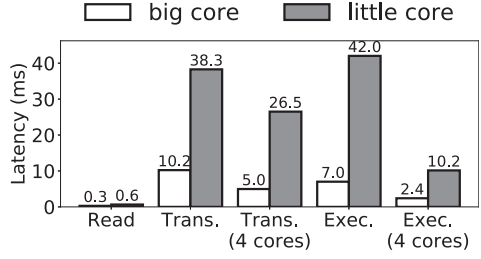

**Figure 6: The consumed time of different stages of cold start on different ARM core types and numbers.**

## 3.3 A Heuristic-Based Kernel Scheduler

---

**Algorithm 1:** Our kernel scheduler

---

**input** : Number of little cores, $M_l$;
Number of layers, $N$ ;
Sets of candidate kernels' combination, $K$;
operations $r_i$, $w_i$, $e_i$ ($i \in \{1, 2, ..., N\}$).

**output** : Combination of selected kernels, $K_c$;
The list of operations running in little core $j$,
$Q_j$ ($j \in \{1, 2, ..., M_l\}$);
The list of operations running in big cores, $Q_0$.

**1** Filter out the kernel candidates that exhibit no faster operation;

**2 foreach** *combination* $k = \{< r_i, w_i, e_i > | i = 1, 2, ...N\}$ $(k \in K)$ **do**

**3**     Initialize $Q_0$: Insert the operations $r_1$, $w_1$ and and $e_1$ of $k$ into the big cores sequentially, s=2;

**4**     Initialize the execution time of the operations on core $j$:
    $T_{Q_j} = 0$, $j \in \{0, 1, ..., M_l\}$;

**5**     Update the execution time of operation $o$ on little cores $t_o^l$ and big cores $t_o^b$;

**6**     **while** $\left| \max\limits_{1 \le j \le M_l} T_{Q_j} - T_{Q_0} \right| > \varepsilon$ or $T_{Q_j} = 0$, $(j \in \{0, 1, ..., M_l\})$ **do**

**7**        **if** $\max\limits_{1 \le j \le M_l} T_{Q_j} > T_{Q_0}$ **then**

**8**           **for** $i \leftarrow s$ to $N$ **do**

**9**              **if** $(t_{r_i}^b + t_{w_i}^b) + (t_{r_i}^l + t_{w_i}^l) < \max\limits_{1 \le j \le M_l} T_{Q_j} - T_{Q_0}$ **then**

**10**                 Insert $r_i$, $w_i$ into $Q_0$ header, $s := i$ ;

**11**                 **break**;

**12**     Initialize $Q_j$ $(j = 1, ..., M_l)$: schedule $r_i$, $w_i$ $(i = s + 1, ..., N)$ to different little cores sequentially;

**13**     **while** $\max\limits_{1 \le j \le M_l} T_{Q_j} - \min\limits_{1 \le j \le M_l} T_{Q_j} > \varepsilon$ **do**

**14**        $j_{max} := \arg\max\limits_{1 \le j \le M_l} T_{Q_j}$

**15**        $j_{min} := \arg\min\limits_{1 \le j \le M_l} T_{Q_j}$

**16**        Sort operations in $Q_{j_{max}}$ descendingly according to the execution time as $Q_{sort}$;

**17**        **foreach** *operation* $(r, w)$ *in* $Q_{sort}$ **do**

**18**           **if** $t_r^l + t_w^l < \frac{T_{Q_{jmax}} - T_{Q_{jmin}}}{2}$ **then**

**19**              Move $(r, w)$ from $Q_{j_{max}}$ to $Q_{j_{min}}$;

**20**     Compute $T_{Q_j}$ $(j = 0, 1, ..., M_l)$;

**21**     Compute the completion time of kernel combination $k$, $T_{cold}^k$ ;

**22** $K_c = \arg\min\limits_k (T_{cold}^k)$

---

**Heuristics** We design our kernel scheduling algorithm based on the following heuristics. First, for almost every DNN we have tested, the kernel execution is still the most time-consuming type of operation. The lower bound we can possibly achieve for cold inference latency is equal to the warm inference, which usually places all the execution operations on big cores with multithreading acceleration. Second, there exists operation-to-hardware affinity. As shown in Figure 6, the big core on Meizu 16T can reduce the execution time by 6× compared to the little core, but can only reduce the weights reading and transformation by 2× and 3.8×, respectively. This is because weights reading and transformation are more likely to be bottlenecked by disk I/O and memory I/O instead of computing. Third, multithreading is more efficient for execution operation than others. Conceptually, every single operation can be multithreaded on multiple cores for acceleration. However, according to results in Figure 6, the speedup of multithreading on kernel execution can almost linearly scale with the number of cores, yet multithreading exhibits poor performance on weights reading and transformation. This is because multithreading is more friendly to computation-intensive operations as it incurs inter-cores synchronization overhead.

**Assumptions** Based on the above heuristics, we build our algorithm atop the following key assumptions.

● *Each kernel's execution* operation *always occupies all big cores and is executed sequentially.*

This is based on our observation that executing execution operation on LITTLE cores could easily bottleneck the whole inference, leaving the big cores under-utilized. Meanwhile, multitasking many execution operations on big cores does not exhibit any improvement as the highly optimized DNN engine could already fully utilize the cores with execution operations. Figure 6 illustrates how kernel execution on big CPU cores achieves the lowest warm inference latency. This is critical to push the performance of cold inference to the limit of warm inference.

● *Weights reading and transformation* operations *of the same operator are always bundled together (as a new preparation* operation*) and mostly placed on little cores without multithreading.* The rationale is that the precursor operation of transformation operations is the weights reading operation of the same operator, which are both I/O intensive and have very few precursor operations (0 or 1) as compared to execution operation (at least 2), therefore can be easily pipelined. Since execution operation occupies all big cores, we can use many little cores to run those operations separately at the same time.

**Algorithm of kernel scheduling.** Our proposed algorithm (Algorithm 1) is composed of two layers. In the outer layer (line 2), we traverse to find the optimal kernel combination. A kernel combination refers to, for each operator, what kernel to use and whether to bypass the weights transformation. There are $\prod_{i=1}^N (2 \cdot c_i)$ such combinations, where $c_i$ is the number of kernel candidates of $i^{th}$ operator. Apparently, we do not need to iterate over all of them; instead, for each operator, we filter out the kernel candidates that exhibit no faster operation in either preparation or execution than any other candidate. After that, there are only 1–2 candidate kernels left for each operator as observed.

In the inner layer, we schedule given kernel combination to minimize the completion time of the last kernel. As each kernel's execution operation always occupies all big cores, we further divide the task to: (1) balance the loads among the little cores to minimize the largest completion time on them; (2) balance the workloads between the little cores and large cores to minimize the completion time of big cores. In the algorithm, we use two loops to solve this problem. In the big-core loop (line 6-11), we determine which operations should be executed on big cores. The operations of the first kernel (to fast boot) and all the execution operations of the rest kernels should run on big cores (line 3). If the completion time of big cores ($T_{Q_0}$) is still less than the largest completion time of the little cores after moving one reading and transformation operation from little cores to big cores (line 9), the weights reading and transformation operation should be inserted to the big cores (line 10); In the little-core loop (line 13-20), the reading and transformation operations are scheduled among the little cores to balance the workloads. We initialize the operation lists of little cores (line 12) by sequentially scheduling the reading and transformation operations one by one to different little cores (as shown in Figure 7(b)). If the little core with the earliest completion time has the potential to accommodate the reading and transformation operations from the little core with the largest completion time (line 18), migrate the reading and transformation operations to balance the workloads (line 19).

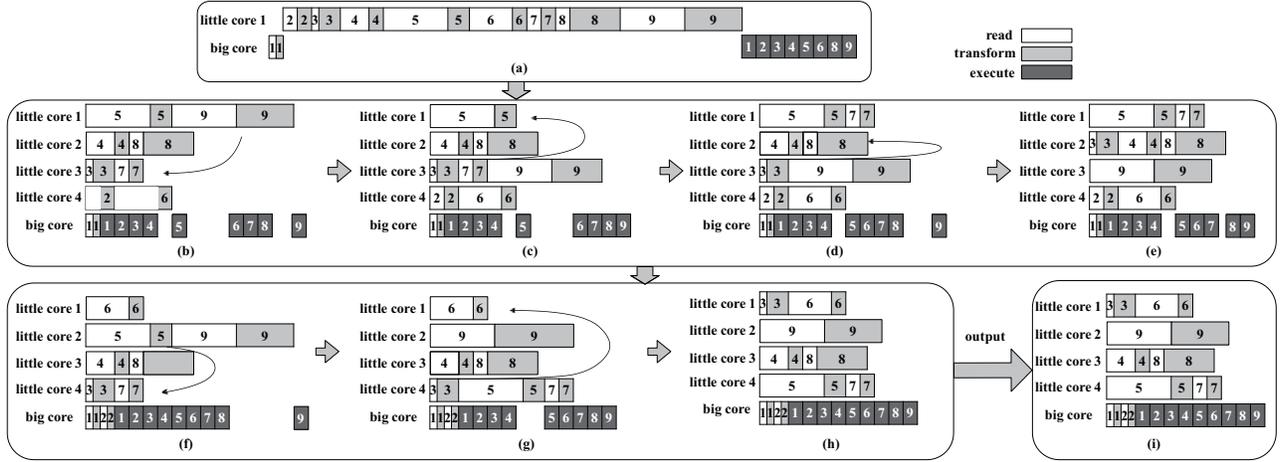

**Figure 7: An illustrative example of how NNV12's kernel scheduling algorithm works (Algorithm 1).**

**An illustrative example** is shown in Figure 7. Figure 7(a) corresponds to Line 3 in Algorithm 1, where we set the reading and transformation operations of layer 1 and all execution operations on big cores, while the other operations are placed on little core. Figure 7(b)–(e) and Figure 7(f)–(h) are two iterations of the big-core loop in Algorithm 1. Figure 7(b)–(e) are four iterations of the little-core loop.

**Dealing with hardware dynamics** NNV12 further introduces a *workload stealing* technique to adapt to hardware dynamics, e.g., cores occupied by other tasks/apps during inference. The key idea is that, once a core is shared by other workloads, the operations scheduled on it will run slower and some of them are better to be relocated to other cores. NNV12 determines such workload stealing on demand: when a busy core slows down the whole inference and another core becomes idle with no other operations to run, that idle core will steal the operations from the head of the job queue of the busy core and execute them accordingly. Such stealing could happen among multiple cores, as long as there are idle cores whose next operation has unfinished dependency.

### 3.4 Extending to CPU/GPU architecture

In the previous sections, we mainly introduce how NNV12 fits big.LITTLE CPU architecture. Conceptually, the above design can be easily extended into GPU platform by treating the GPU as the big core and CPU as little cores. Yet, the unique characteristics of GPUs require NNV12 to make further revisions and optimizations to achieve optimal performance.

**Creating pipeline as another `operation`** For each operator, in addition to the weights reading, transformation, and kernel execution on CPU, there is another `operation` in the cold inference namely *creating pipeline* [11]. Taking Vulkan as an instance, this step sets up a pipeline that describes the configurable state of the graphics card, like the viewport size and depth buffer operation. It is usually implemented with ahead-of-time compilation [12], therefore incurs no overhead for warm inference. In cold inference, however, this operation can take a considerable amount of time to run as previously shown in Table 1.

**Operations-to-processor placement** The GPU is only in charge of kernel execution while all other operations are scheduled on CPUs as latter can hardly be accelerated by GPU. It also helps reduce the CPU-GPU data copy. Further partitioning the execution across CPU and GPU [34] might enlarge the optimization spaces but is orthogonal to this work and left to be explored in the future.

**Caching compute shaders** One time-consuming and GPU-specific process we observed is shaders compiling [6]. In Neural Networks, a kernel is implemented as a shader [14]. For example, 3D graphics and compute API Vulkan drivers are supposed to ingest shaders already translated into an intermediate binary format called SPIR-V (Standard Portable Intermediate Representation). For a given DNN model, the shaders that need to be compiled and generated at each layer are determined. Therefore, we can cache those shaders on disk and load them directly to speed up the cold inference just as how we bypass the weights transformation stage.

### 3.5 Kernel Switching for Warm Inference

The kernels selected by NNV12 are optimized for cold inference. As discussed in §3.1, the kernels with the fastest warm inference might be different from what NNV12 selects. We use $K_{cold}$ and $K_{warm}$ to represent two different sets of kernels. If NNV12 keeps using the kernels of $K_{cold}$ in subsequent inferences, it leads to a suboptimal warm inference latency.

| Model | Task | Parameters | Model Size | FLOPs | Storage Overhead | Scheduling Plan Generation Time (Offline) | | | |
|---|---|---|---|---|---|---|---|---|---|
| | | | | | | Meizu 16T | Pixel 5 | TX2 | Nano |
| AlexNet [33] | C | 61.3M | 237.5M | 1.4G | 172.3M | 1,538.4ms | 4,157.4ms | 8,197.3ms | 22,962.6ms |
| GoogLeNet [53] | C | 7.1M | 26.9M | 3.2G | 22.6M | 1,301.9ms | 2,304.3ms | 6,457.6ms | 9,648.7ms |
| MobileNet [27] | C | 4.4M | 16.2M | 1.1G | 12.6M | 796.5ms | 1,796.5ms | 5,443.0ms | 8,357.5ms |
| MobileNetV2 [52] | C | 3.7M | 13.3M | 0.6G | 10.3M | 759.3ms | 1,796.5ms | 4,770.0ms | 8,441.1ms |
| ResNet18 [26] | C | 12.7M | 45.5M | 3.9G | 34.3M | 892.6ms | 1,896.2ms | 1,461.1ms | 2,599.1ms |
| ShuffleNet [73] | C | 3.6M | 12.9M | 1.9G | 10.0M | 577.8ms | 1,005.9ms | 5,872.3ms | 7,996.8ms |
| EfficientNetB0 [54] | C | 5.4M | 19.6M | 0.8G | 15.2M | 1,129.9ms | 2,446.0ms | 6,481.2ms | 6,031.4ms |
| ResNet50 [26] | C | 25.7M | 97.4M | 7.8G | 89.7M | 1,652.2ms | 2,974.3ms | 3,757.6ms | 3,854.0ms |
| SqueezeNet [29] | C | 1.4M | 4.7M | 1.7G | 3.8M | 717.9ms | 1,788.8ms | 5,849.8ms | 6,738.9ms |
| ShuffleNetV2 [42] | C | 3.4M | 12.0M | 0.5G | 10.9M | 532.1ms | 920.7ms | 4,724.5ms | 5,665.3ms |
| MobileNetv2-YOLOv3 [51] | OD | 3.6M | 13.1M | 1.0G | 12.5M | 849.2ms | 2,544.5ms | 3,394.3ms | 4,979.7ms |
| MobileNet-YOLO [50] | OD | 11.9M | 49.1M | 2.9G | 38.3M | 984.2ms | 2,485.5ms | 5,047.5ms | 7,710.2ms |
| CRNN-lite [24] | OCR | 2.4M | 2.6M | 3.1G | 45.4M | 116.32ms | 375.32 ms | 4,597.6ms | 6,257.3ms |

**Table 4: The NN models used in experiments. Input size: 224x224. "C": classification; "OD": Object Detection; "OCR": Optical Character Recognition.**

To handle such side effects, NNV12 provides an additional mode besides the one only optimized for cold inference as discussed above. This mode indicates that there will be continuous inferences tasks. In that case, NNV12 still follows the aforementioned techniques to optimize the cold inference, but makes the following key differences: (1) It also prepares the kernels in $K_{cold} - K_{warm}$ and switches to kernels in $K_{warm}$ for later inferences. (2) The preparation of those additional kernels is performed on little cores when idle during the cold inference. The rationale is that the little cores have some idle time before the kernel execution finishes on big cores. If such idle time is not enough to prepare the kernels in $K_{cold} - K_{warm}$, the rest of the `operations` will be pipelined in the second inference as NNV12 does for the cold inference. In §4.5 we experimentally show that NNV12 achieves (near-)optimal performance in continuous inference as well.

## 4 EVALUATION

### 4.1 Implementation and Methodology

**NNV12 prototype** We've implemented a prototype of NNV12 with 18K C++ LoC atop ncnn (version 20211208) for its lightweight codebase and superior performance as compared to TFLite. We used Vulkan GPU backend for its more generic support for different platforms. Note that the techniques of NNV12 are compatible with other DL libraries as well.

**Models** We use 12 popular NN models as summarized in Table 4 to test the performance of NNV12. Those models span across different tasks (image classification and object detection) and computation complexity. We mainly use CNN models in our experiments because a recent empirical study shows that CNNs are dominant use cases in nowaday edge devices [13]. The models mainly come from the official model zoo of those libraries [9], while for the ones that do not exist in the zoo, we generate them by ourselves, e.g., implementing the model structure in TF APIs and then converting it to TFLite format. We manually check that the same model used by each library has a consistent structure.

**Hardware** We use 6 different devices: Meizu 16T smartphone with Snapdragon 855, Google Pixel 5 with Snapdragon 765G, Redmi 9 with MTK Helio G80, Meizu 18 Pro with Snapdragon 888, Jetson

TX2, and Jetson Nano. The OS of Meizu 16T and Google Pixel 5 is Android 11. The OS of Jetson TX2 and Jetson Nano is Ubuntu 18.04. We use only CPUs for the two smartphones and use GPUs on the Jetson boards. The reason is that, on smartphone SoCs, the CPUs perform much better than GPUs for cold inference as DNN preparation takes much more time than CPU as shown in Table 1. Yet, on Jetson TX2/Nano with much more powerful GPUs, the DNNs are almost always placed on GPUs.

**Baselines** On Meizu 16T and Pixel 5, we compare the performance of NNV12 to 3 baselines: ncnn, TFLite, and AsyMo [56]. Since NNV12 is implemented atop ncnn, the comparison between them can directly reveal the effectiveness of NNV12's key techniques. Still, TFLite is added as it is the most popular DL library used in end devices (version 2.5.0). AsyMo is the state-of-the-art DL engine that can fully exploit the asymmetric CPU architecture on smartphones. Since AsyMo is not open-sourced yet, we re-implement it atop ncnn for a fair comparison. On Jetson TX2/Nano, we also use ncnn with its Vulkan backend. However, since TFLite does not support either Vulkan or CUDA backend, we replace it with TensorFlow (version 2.5.0) for comparison.

**Setups and configurations** On Meizu 16T and Pixel 5, we exhaustively try different core numbers for TFLite and ncnn and use the best configuration. In practice, it turns out to be 4 cores on Meizu 16T and 2 cores on Pixel 5. Note that AsyMo always uses all the CPU cores available. The model files are stored on SDCards for both smartphones and Jetson boards. To eliminate the impacts of file cache, we clear the system cache before each cold inference. For all experiments, the cold inference latency does not include the loading and initialization time of the libraries. Each experiment is repeated by 100 times and the average numbers are reported.

### 4.2 End-to-End Performance

**Cold inference latency on CPU** Figure 8 compares the cold inference latency of NNV12 with the baselines on edge CPUs and Table 5 summarizes NNV12's overall improvements. It shows that NNV12 significantly outperforms the baselines on all models and platforms, i.e., 1.1×–15.2× speedup over TFLite and 1.2×–10.3× speedup over ncnn.

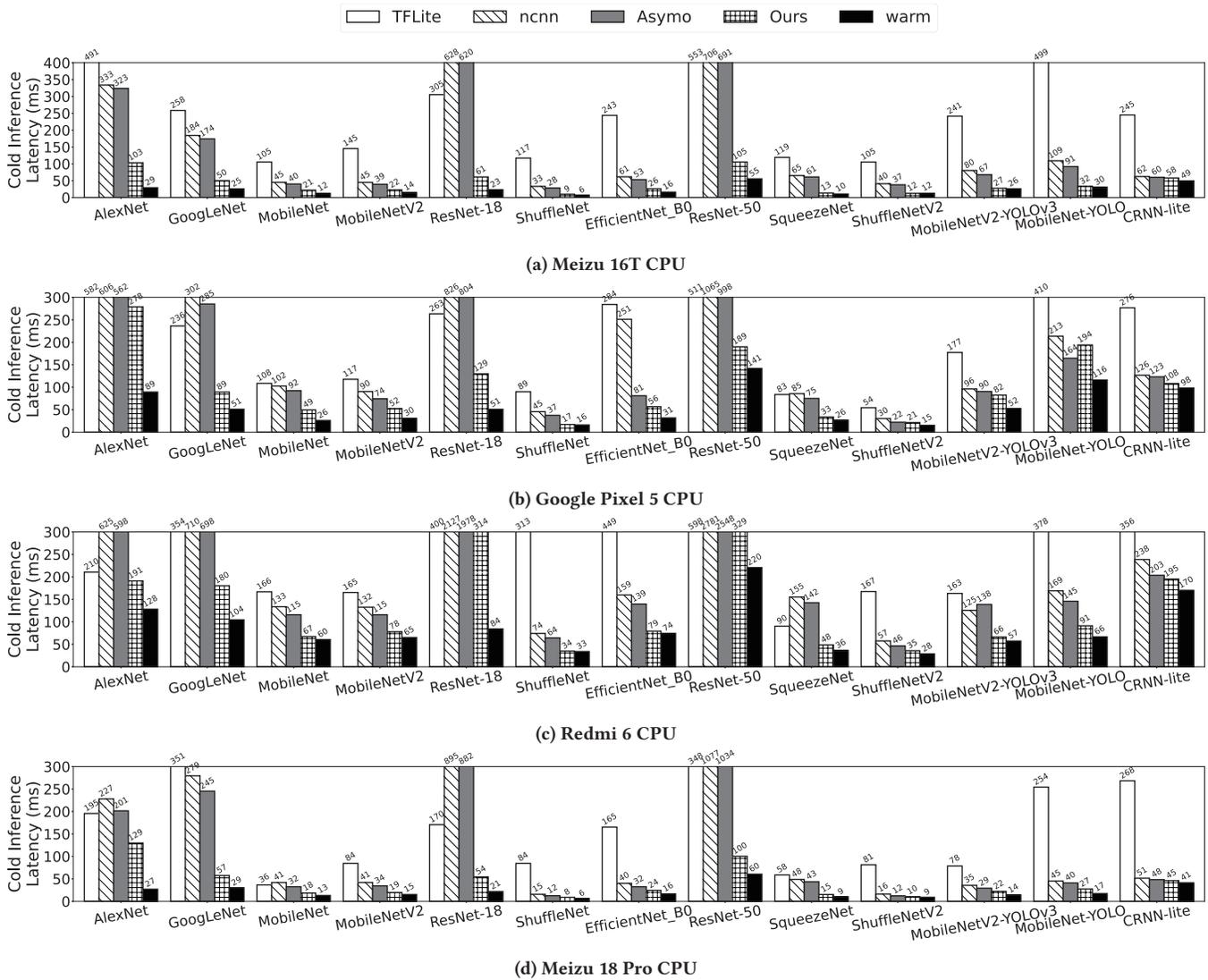

**Figure 8: The cold inference latency of NNV12 and baselines on edge CPUs.**

**Table 5: Summarized performance comparision of NNV12 over baselines on different platforms.**

| HW Platform | Speedup over baselines (min − max, avg) | |
|---|---|---|
| | ncnn | TFLite (TF) |
| Meizu 16T (CPU) | $1.1\times - 10.3\times$ (3.7×) | $4.2\times - 15.2\times$ (7.5×) |
| Pixel 5 (CPU) | $1.1\times - 6.4\times$ (2.8×) | $2.1\times - 5.2\times$ (2.2×) |
| Redmi 9 (CPU) | $1.2\times - 8.5\times$ (3.1×) | $1.1\times - 8.9\times$ (3.2×) |
| Meizu 18 Pro (CPU) | $1.2\times - 16.4\times$ (3.9×) | $1.5\times - 9.4\times$ (5.2×) |
| Jetson TX2 (GPU) | $9.0\times - 38.9\times$ (29.6×) | $14.6\times - 355.3\times$ (154.8×) |
| Jetson Nano (GPU) | $4.0\times - 58.2\times$ (28.5×) | $10.4\times - 401.5\times$ (234.3×) |

NNV12 also achieves close performance to warm inference, i.e., only 1.72× slower at average. On ShuffleNetV2, the gap is even negligible (≤1ms). This is because NNV12 effectively overlaps the preparation stages (loading and transformation) with the execution, therefore their latency can be mostly hidden. Yet, the gap still exists for three reasons. First, the model execution could be waiting for the preparation to be done on CPU little cores when the overlapping is not perfectly planned. Second, even without waiting, the execution could be slowed down due to the cross-operation interference as mentioned in §3.3. Third, NNV12 selects kernel for fast cold inference, whose real execution speed might be slower than the original selection that optimizes for the warm inference.

The more competitive baseline AsyMo achieves relatively limited improvements over the vanilla DNN engine ncnn, i.e., only 1.03×– 1.28× speedup. This is because it only improves the execution speed by fully utilizing the asymmetric CPU cores through kernel scheduling, but the weights preparation still takes a considerable amount of time in cold inference.

**Impacts of CPU core numbers** In the above experiments, we always set the CPU core number to the one obtaining the best performance for TFLite and ncnn. In practice, it turns out to be 4 on Meizu 16T and 2 on Pixel 5. Figure 9 further shows a comprehensive comparison by using different core numbers on Meizu 16T. It confirms our observation that using 4 cores exhibits the best performance for TFLite and ncnn, which is also consistent with the prior study [56]. This is mainly because those DL engines cannot well exploit the asymmetric CPU cores for DL execution. Note that they also use multi-threads to accelerate the weights transformation yet the profit from more threads is also marginal. This is because the weights transformation is mainly memory-bounded. Instead, NNV12 pipelines different kinds of operations to fully exploit the disk, memory, and computing capacity.

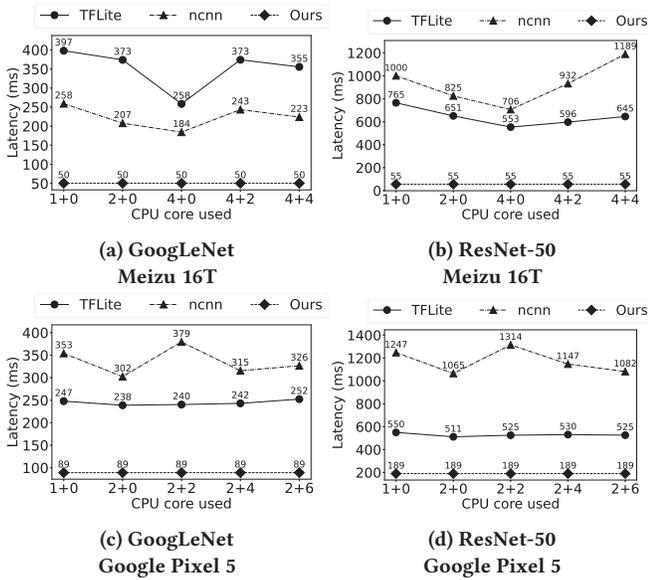

Figure 9: The cold inference latency of NNV12 and baselines running on different CPU core numbers. "X+Y" indicate X big cores and Y little cores.

**Cold inference latency on GPU** Figure 10 compares the cold inference latency of NNV12 with the baselines on edge GPUs and Table 5 summarizes NNV12's overall improvements. It shows that the performance improvement of NNV12 compared to the baselines is even more significant than CPU, i.e., 10.4×–401.5× speedup over TensorFlow and 4.0×–58.2× speedup over ncnn. There are two primary reasons for such a huge improvement. First, the cold inference on GPU requires more preparations such as creating pipeline. Therefore NNV12's key techniques, especially the kernel pipeline (§3.3), bring more benefits. Second, NNV12 incorporates additional optimizations for GPU like shader caching (§3.4).

**Dynamic Loads** We also evaluate how NNV12 adapts to the dynamic background loads as compared to vanilla ncnn. We use a customized program to impose different levels of pressure (0%/25%/50% CPU utilization) on different cores in background while the cold inference takes place. Figure 11 shows the testing results with GoogleNet on the Meizu 16T device. It shows that, when little cores are occupied and if NNV12 sticks to the optimal kernel scheduling plan generated offline, its performance degrades significantly, e.g., up to 2.1×. This is because NNV12 schedules the read and transformation operations across the little cores; when some of them are busy, they bottleneck the whole cold inference process. Meanwhile, the performance of ncnn is not affected as it only leverages the 4 big cores to obtain the best possible performance. Nevertheless, thanks to the workloads stealing technique, NNV12 does not bottleneck on the little cores but make dynamic decision to balance the workloads across all cores. With 2 little cores occupied by 50% each, NNV12's cold inference performance only drops from 50ms to 75ms and is still 2.5× faster than ncnn. On the other hand, when big cores are occupied, the performance of NNV12 degrades more significantly as well as vanilla ncnn.

**Energy consumption** We also evaluate the energy consumption of NNV12 and illustrate the results in Figure 12. We observe that NNV12 can significantly reduce the energy consumption, i.e., 0.2×–0.6× compared to ncnn. Such energy-saving mainly comes from the saved inference time through NNV12's key techniques, especially the kernel selection.

### 4.3 Ablation Study

We then evaluate the benefits brought by NNV12's each key technique separately. The results are illustrated in Figure 13. Our major observation is that each of NNV12's key techniques contributes noticeably to the cold inference speedup. For example, with ResNet-50 and Jetson TX2, the kernel selection first reduces the cold inference latency from 8,272ms to 2,300ms. Caching the post-transformed weights further reduces the latency to 555ms, and with pipelined execution the latency finally becomes 240ms.

### 4.4 Resource Overhead

There are two kinds of overhead NNV12 introduces: at offline, NNV12 needs to generate the optimal kernel scheduling plan according to (§3.3); to boost the cold inference, NNV12 opportunistically stores the post-transformed weights on disk in addition to the raw weights (§3.1). (1) **Time to generate scheduling plan** As shown in Table 4, NNV12 takes only 532.1ms–4157.4ms on Meizu 16T and Pixel 5 CPU to generate the kernel scheduling plan. It takes more time on Jetson TX2 and Nano, i.e., 1461.1ms–22962.6ms, because of the more complicated preparation stages of GPUs. Note that this overhead only occurs for one shot when a model is fetched to a device, and shall not compromise the user experience. (2) **Disk storage for post-transformed weights** As shown in Table 4, the storage overhead to cache the post-transformed weights is 7.1MB–164.8MB. Note that not every layer will apply the cache technique depending on the operator characteristics and kernel scheduling strategy. Since nowaday edge devices are typically equipped with a few to tens of GBs disk, such storage overhead is tolerable.

### 4.5 NNV12 in Continuous Inference

Recall that NNV12 incorporates a unique design (§3.5) to optimize for consecutive inferences as well. We experiment with GoogleNet and ResNet-50 on Meizu 16T. The results are illustrated in Figure(14). It shows that NNV12 not only greatly optimizes the cold inference latency, but also achieves close performance to ncnn in the second inference, i.e., only 8% slower, and the same speed since the 3rd inference. NNV12 runs slightly slower on the second inference than ncnn because the idle time of little cores during cold inference might not be enough to prepare all the kernels for the warm inference. In that case, NNV12 follows the pipeline design to speed up the second inference.

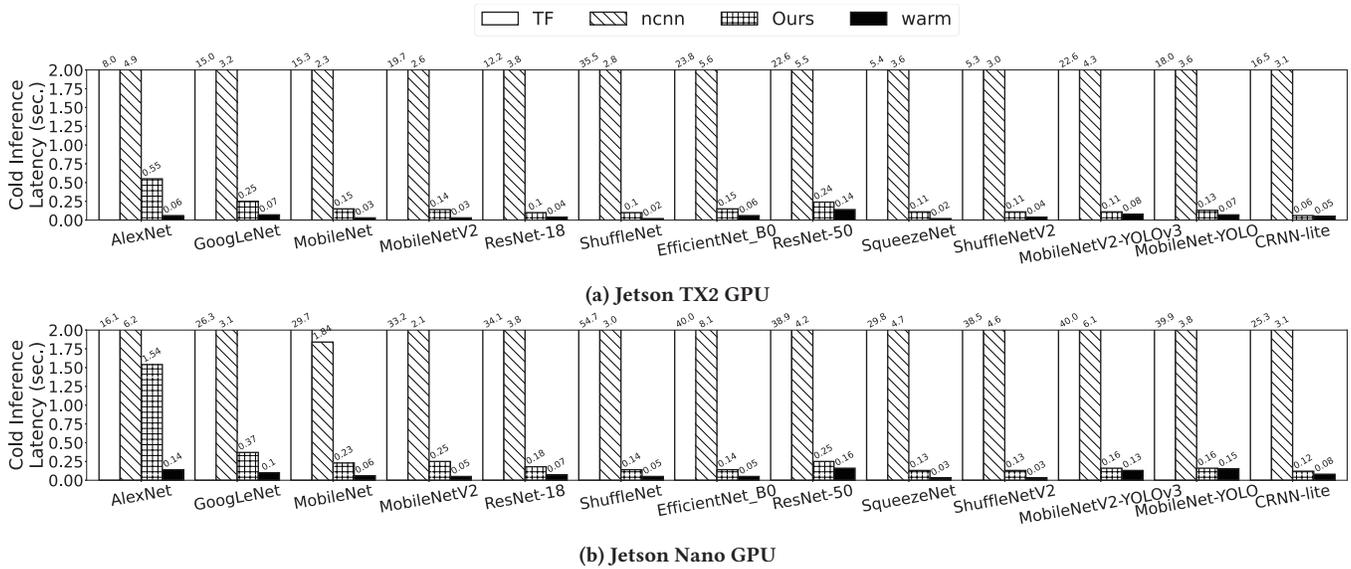

Figure 10: The cold inference latency of NNV12 and baselines on edge GPUs.

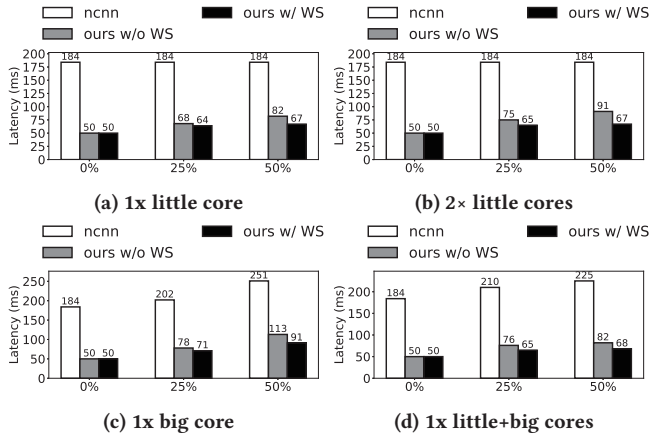

Figure 11: The performance of NNV12 adapting to dynamic background workloads. The numbers 0%/25%/50% indicate the background load on the CPU cores. "WS": workloads stealing technique.

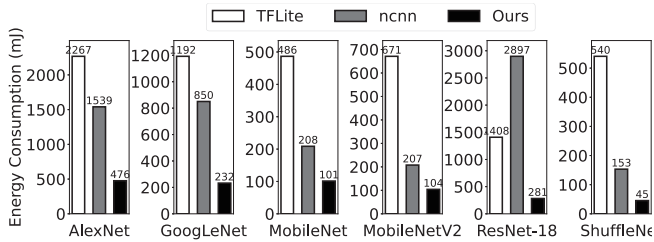

Figure 12: The energy consumption of cold inference.

## 5 RELATED WORK

**DNN weights sharing** To reduce the memory footprint of multiple concurrent DNNs, prior works [23, 34, 38, 56, 59, 67, 72] have proposed to let the DNNs share certain layers of weights (especially the beginning ones). This approach has the scalability issue as with more DNNs the model accuracy can drop significantly. Or, they virtualize the DNN weights memory to better manage the data in/out switching among DRAM and disk [38]. This approach still incurs a high overhead in data swapping, thus does not address the slow cold start inference. Instead, this work directly optimizes the cold inference and does not compromise accuracy.

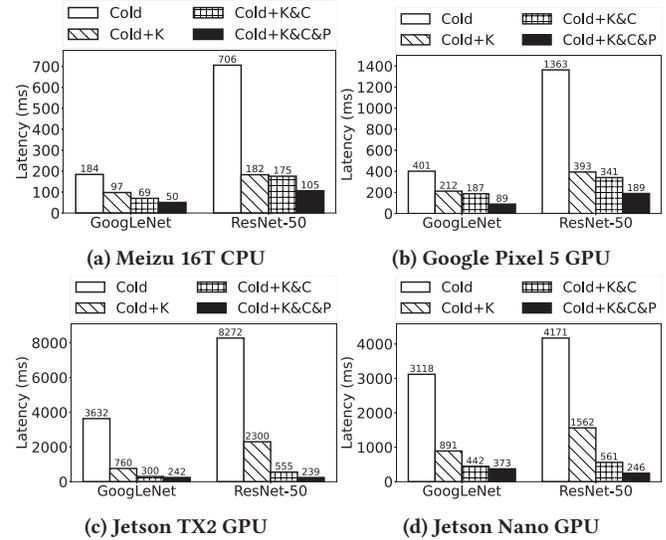

Figure 13: The ablation study results of NNV12. "K": kernel selection; "C": caching the post-transformed weights (and shaders); "P": kernel execution pipeline.

**Apps pre-launch** Mobile apps also face the cold launch problem. Prior works mainly use pre-launching [15, 48, 68] to mitigate the slow cold launch problem, i.e., by predicting when an app is going to be launched soon so the OS can prepare it. Intuitively, we can retrofit

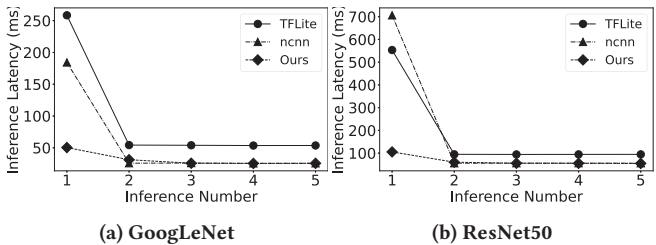

**(a) GoogLeNet**

**(b) ResNet50**

**Figure 14: The cold inference and subsequent warm inference latency of NNV12 and baselines.**

this idea to reduce the cold inference times of DNNs as well. Yet, it has the following drawbacks. First, there will be much more DNNs than apps on a device [13, 63], making an accurate prediction difficult. Second, unlike apps, DNNs are transparent to mobile OSes, thus there is no unified interface for OSes to bookkeep and operate on the DNNs hosted on a device.

**DNN fast switch on clouds** PipeSwitch [16] enables fast switch among training and inference tasks on the same cloud GPU. It inspired some of NNV12's design points, e.g., pipelined I/O and execution by exploiting the layer-by-layer structure of DNNs. However, PipeSwitch is not designed for cold inference optimization, as it does not consider the model loading and weights transformation stages. Therefore it's not directly comparable to NNV12.

**DNN inference optimizations** There are two main categories of on-device DNN inference optimizations. One is at system level, e.g., by exploiting heterogeneous processors [20, 25, 28, 34], cache [45, 61, 67], generating high-performance GPUs kernels [40], or adaptive offloading [35, 65]. Such methods only work for warm inference. The other one is model level, e.g., quantization [32, 41] or sparsification [17, 46]. While those works mainly target warm inference, conceptually, they can also improve the cold inference as they reduce the execution time and/or the weights to be read from disk. NNV12 explores optimization knobs from different aspects and is orthogonal to them.

## 6 CONCLUSIONS

In this work, we propose NNV12, the first DL engine that optimizes the cold inference on edge devices. NNV12 fully exploits three optimizations knobs: kernel selection, weights transformation caching, and pipelined inference through a heuristic-based kernel scheduling scheme. Extensive experiments demonstrate the effectiveness of NNV12 to boost cold inference on edge CPU and GPU hardware.

## 7 ACKNOWLEDGEMENT

This work was supported by National Key R&D Program of China (No.2021ZD0113001), NSFC (No. 62102045, No.62032003, No.U21B2016), Beijing Nova Program (No.Z211100002121118), and Young Elite Scientists Sponsorship Program by CAST (No.2021QNRC001). We thank the anonymous shepherd and MobiSys reviewers for their valuable suggestions.

## A ARTIFACT APPENDIX

The research artifact accompanying this paper is available via https://doi.org/10.5281/zenodo.7922815.